\documentclass[12pt, a4paper]{article}

\usepackage[margin=1in]{geometry}

\usepackage[T1]{fontenc}
\usepackage{rotating}
\usepackage{graphicx}
\usepackage{amsmath,amssymb}
\usepackage{booktabs}
\usepackage{multirow}
\usepackage{multicol}
\usepackage{subcaption}
\usepackage{tikz}
\usepackage{fontawesome5}
\usepackage{arydshln}

\usetikzlibrary{arrows.meta, positioning}
\usepackage{authblk} 

\usepackage{hyperref}
\hypersetup{
    colorlinks,
    linkcolor={red!50!black},
    citecolor={blue!50!black},
    urlcolor={green!50!black}
}

\begin{document}

\title{{GRC-ProbNet}: Uncertainty-aware Feature Extraction for Cardiovascular Disease Classification}

\author[1]{Yash Shah}
\author[1]{Omar Todd}
\author[2]{Philipp Seeböck}
\author[2]{Georg Langs}
\author[1]{Ben Glocker}
\author[1]{Raghav Mehta\textsuperscript{\faEnvelope[regular], }}

\affil[1]{\small Imperial College London, UK}
\affil[2]{\small Medical University of Vienna, Austria}

\affil[ ]{}
\affil[ ]{\small \textsuperscript{\faEnvelope[regular]} Email: raghav.mehta@imperial.ac.uk}

\date{} 

\maketitle

\begin{abstract}
The automatic detection and classification of cardiovascular disease (CVD) from computed tomography (CT) images plays an important role in clinical practice. Recently, a hybrid pipeline (\texttt{GRC-Net}) for CVD classification was proposed, which leverages a deep-learning-based segmentation and registration method to extract radiomic and geometric features. However, \texttt{GRC-Net} relies on a deterministic segmentation mask, without considering the inherent ambiguity associated with cardiac anatomy. In this paper, we propose \texttt{GRC-ProbNet}, which takes advantage of a deep ensemble to produce multiple segmentation masks for a given input. From these masks, we extract multiple uncertainty features. We analyze these uncertainty features for both their correlation with segmentation error and their propagation effects on downstream CVD classification performance. Our experiments on the publicly available MM-WHS and ASOCA datasets show that the uncertainty measure that best reflects segmentation quality is not necessarily the one that provides the strongest signal for downstream CVD classification. Overall, our results demonstrate that \texttt{GRC-ProbNet} utilizing uncertainty features substantially improves CVD classification AUROC (92.92\%) compared to the baseline \texttt{GRC-Net} model (91.25\%). Our code is publicly available: \url{https://github.com/biomedia-mira/GRC-ProbNet}.

\vspace{0.5cm}
\end{abstract}

\section{Introduction}

Cardiovascular disease (CVD) remains one of the leading causes of mortality worldwide, accounting for an estimated 20.5 million deaths in 2025, with cardiovascular mortality projected to rise by 73.4\% between 2025 and 2050 as populations age~\cite{Chong2025}. Cardiac computed tomography (CT) has become a central non-invasive imaging modality for the diagnosis of CVD. However, analysing these scans manually is time-consuming, requiring significant clinical expertise~\cite{luijnenburg2010intra}. These challenges motivate the development of automated, segmentation-based pipelines that can deliver fast, consistent, and reproducible analysis at scale.  

Deep learning is the leading approach for analysing cardiac images. It has been extensively applied to the automated detection and quantification of coronary artery disease in CT angiography, demonstrating strong agreement with expert radiologist annotations~\cite{lin2022,litjens2017,alven2025}. In addition to improving accuracy, these methods often reduce reporting times and support decision-making in time-critical settings such as acute cardiac care~\cite{topol2019}. A key limitation of existing cardiac image analysis pipelines is their inability to accurately represent the inherent ambiguity observed in cardiac anatomy. Their deterministic outputs do not align with the inter-rater variation that would be present across multiple expert annotations. This limitation reduces their efficacy for meaningful uncertainty propagation and analysis in downstream stages.

Recently, many methods have been proposed that allow approximating the uncertainty associated with model outputs; these include approximate Bayesian methods, such as Monte Carlo Dropout~\cite{gal2016dropout}, Deep Ensembles~\cite{lakshminarayanan2017}, and Stochastic Segmentation networks~\cite{monteiro2020}. 
Each of these produces multiple plausible predictions, and their variability quantifies model uncertainties. 
Recent works report that propagating these uncertainties in cascaded inference tasks leads to improvement in downstream tasks~\cite{mehta2021,feiner2023,Wundram2024,Fischer2023}. However, most of these existing approaches propagate uncertainty directly between deep networks, bypassing explicit feature extraction. Consequently, a key question remains unanswered: when intermediate explicit features are extracted between cascaded tasks, what is the most effective way to represent and propagate this uncertainty?

This motivates two research questions. First, can segmentation uncertainty provide a complementary diagnostic signal for downstream CVD classification within a hybrid pipeline with both deep learning and non-deep learning components? Second, which representation and encoding of uncertainty should be propagated for downstream classification tasks, and in particular, does the uncertainty that best predicts segmentation error also improve classification performance the most? We make the following contributions:
\begin{itemize}
  \item We introduce \texttt{GRC-ProbNet}, an uncertainty-aware framework that propagates several uncertainty representations -- including predictive entropy, KL divergence, and aleatoric maps -- through an atlas-based radiomic and geometric feature extraction pipeline for CVD classification.
  \item We systematically evaluate different uncertainty types (total, epistemic, and aleatoric) and encoding strategies (global versus per-structure), comparing them against a deterministic baseline.
  \item We find that the uncertainty measure most correlated with segmentation error is not necessarily the one that performs best for classification. Predictive entropy aligns most closely with segmentation error, whereas KL divergence provides more consistent downstream performance improvement.
\end{itemize}

\begin{figure}[t]
\centering
\includegraphics[width=\textwidth]{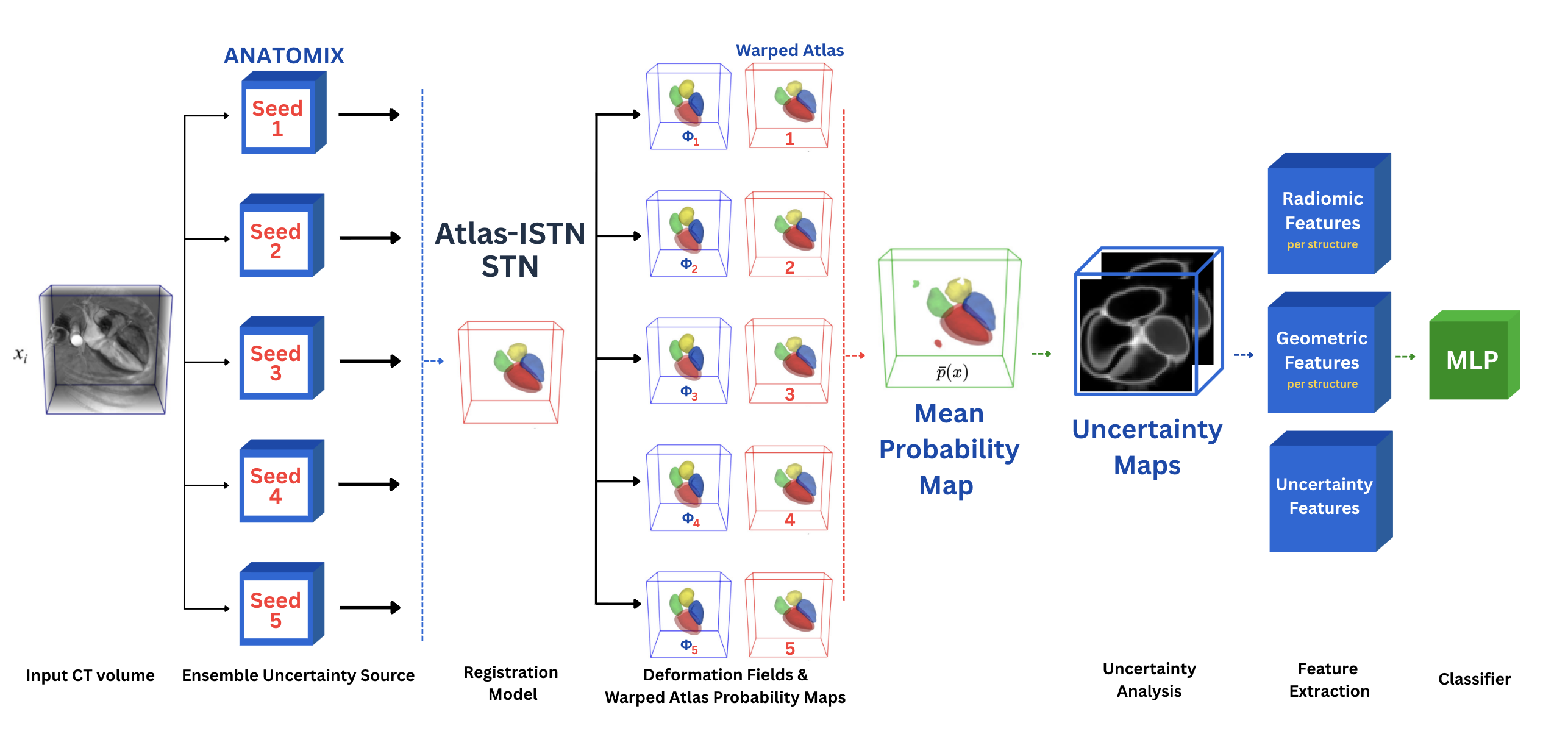}
\caption{Overview of the proposed framework. $N$ (ex., 5) independently fine-tuned Anatomix models segment each input CT; outputs pass through an Atlas-ISTN, producing deformation fields and warped atlas probability maps. These are aggregated into a mean probability map, from which the consensus segmentation and the uncertainty maps are derived. From these, the radiomic, geometric, and uncertainty features are extracted and propagated to a multilayer perceptron (MLP) for downstream CVD classification.}
\label{fig:multiseed}
\end{figure}

\section{Method}

We build on \texttt{GRC-Net}~\cite{mittal2025}: a modular hybrid pipeline that decomposes CVD classification into a sequence of interpretable stages, and extend it with an explicit uncertainty stage, producing \texttt{GRC-ProbNet}. The resulting pipeline consists of four stages (Fig.~\ref{fig:multiseed}): (i) segmentation of cardiac structures from the input CT; (ii) registration to a population-derived healthy atlas, producing deformation fields; (iii) uncertainty maps to quantify segmentation variability; and (iv) downstream CVD classification using radiomic, geometric, and uncertainty features. \\

\noindent\textbf{Segmentation and Registration:} Following GRC-Net~\cite{mittal2025}, we leverage the Anatomix model~\cite{anatomix_web} for the segmentation component and the Atlas Image-and-Spatial Transformer Network (Atlas-ISTN) framework~\cite{sinclair2020} for registration and atlas construction. Anatomix has been shown to outperform other cardiac structure segmentation models~\cite{mittal2025}, such as TotalSegmentator~\cite{wasserthal2023}. For our pipeline, we fine-tune the Anatomix model on a small dataset (see Sec.~\ref{sec:datasets}). The Atlas-ISTN framework jointly learns segmentation and registration while constructing a population-based anatomical atlas. This atlas captures statistical anatomical variations and enables the extraction of geometric features, such as deformation fields, which quantify deviations from the normative atlas. To mitigate Atlas-ISTN's reliance on large, labeled datasets for segmentation, we integrate our fine-tuned Anatomix model into the framework. Finally, we train the registration and atlas construction components of Atlas-ISTN on the same dataset. \\

\noindent\textbf{Radiomic and Geometric Feature Extraction:} We extract two complementary feature types per cardiac structure based on the segmentation masks and atlas registrations obtained from the previously described models. Radiomic features, which characterise tissue appearance and morphology, are computed from the CT intensities within the Anatomix-generated structure segmentation masks. Using PyRadiomics~\cite{vanGriethuysen2017}, we extract first-order statistics describing the intensity distribution (e.g., mean, variance, skewness), 3D shape descriptors (e.g., volume, surface area, sphericity), and texture features that describe the spatial arrangement of intensities. This produces 107 radiomic features per structure. Additionally, geometric features describe how each structure deforms relative to the healthy atlas. Specifically, per-structure displacement vectors are taken from the deformation field output by Atlas-ISTN and reduced via principal component analysis (PCA). This results in up to three eigenvalues per structure, which serve as compact descriptors of anatomical variation. Finally, for a given number of cardiac structures ($S$), the radiomic and geometric features are concatenated into a single combined vector: $\underbrace{(107 \times S)}_{\text{radiomic}} \;+\;
\underbrace{(3 \times S)}_{\text{geometric}} \;=\; 110 \times S \;\text{features}.$

\subsection{Uncertainty Estimation and Feature Extraction}
We estimate uncertainty through a multi-seed deep ensemble~\cite{lakshminarayanan2017}, as it has been shown to perform better than other uncertainty generation methods in previous studies~\cite{mehta2021,Wundram2024}. As shown in Fig.~\ref{fig:multiseed}, five Anatomix models are independently fine-tuned under identical training configurations, differing only in the initial random seed. The disagreement between them serves as an approximation of segmentation uncertainty. All five segmentations are passed through the \emph{same} frozen Atlas-ISTN. The choice of five seeds is a practical trade-off between computational cost and variability coverage.  

\subsubsection{Uncertainty Representations:}
Predictive uncertainty decomposes into two components: \emph{aleatoric} and \emph{epistemic}~\cite{kendall2017}. Aleatoric uncertainty arises from noise and ambiguity that is inherent within the data. Epistemic uncertainty reflects limited knowledge in the model. Both are captured by the deep ensemble~\cite{Kahl2024}.

Let $K$ be the number of seeds, $C$ the number of classes, $\mathbf{x}$ a voxel, and $p^{(i)}_c(\mathbf{x})$ the softmax probability assigned to class $c$ by seed $i$. Averaging the per-seed predictions gives a mean probability: $\bar{p}_c(\mathbf{x}) = \frac{1}{K}\sum_{i=1}^{K} p^{(i)}_c(\mathbf{x})$. Using this value for each class, we can calculate the \emph{total} predictive entropy, which captures both the epistemic and aleatoric components: $$\underbrace{H\!\left[\bar{p}(\mathbf{x})\right]}_{\text{total}} = -\sum_{c=1}^{C} \bar{p}_c(\mathbf{x}) \log \bar{p}_c(\mathbf{x}).$$ 

Epistemic disagreement is described by the divergence of each seed from the consensus, and hence the Kullback--Leibler (KL) divergence is used to measure how far a single seed deviates from the mean map. Averaging the KL across seeds gives the Jensen-Shannon divergence (JSD), which is a summary of inter-seed disagreement~\cite{Kahl2024}: $$\mathrm{KL}\!\left(p^{(i)}(x) \,\|\, \bar{p}(x)\right)
= \sum_{c} p^{(i)}_c(x)
\log\frac{p^{(i)}_c(x)}{\bar{p}_c(x)},$$

$$\underbrace{\mathrm{JSD}\left[\bar{p}(\mathbf{x})\right]}_{\text{epistemic}}
=
\frac{1}{K}\sum_{i=1}^{K}
\mathrm{KL}\!\left(p^{(i)}(x) \,\|\, \bar{p}(x)\right).
$$

Subtracting this epistemic term from the total entropy isolates the \emph{aleatoric} component~\cite{Kahl2024}. This is the uncertainty that remains within the individual predictions once the inter-seed disagreement is removed: $$\underbrace{\mathrm{AE}\left[\bar{p}(\mathbf{x})\right]}_{\text{aleatoric}} = H\!\left[\bar{p}(\mathbf{x})\right] - \mathrm{JSD}\left[\bar{p}(\mathbf{x})\right].$$

These values are calculated at the voxel level to produce either a single global uncertainty map or $C$ structure-specific maps, where $C$ represents the total number of segmented structures. From these maps, we extract image-level uncertainty features by averaging the values in two manner: (i) across all voxels from global uncertainty maps, or (ii) exclusively within each assigned structure from structure-specific maps. While the single-feature approach captures the global uncertainty of an image, the $C$-feature approach provides structure-specific uncertainty. We compare these maps/feature representations for each uncertainty measure against the segmentation error map (the difference between ground-truth and predicted segmentations) before propagating the features to the downstream classifier.

\section{Experiments and Results}

\subsection{Datasets and Implementation Details}\label{sec:datasets}
We use two cardiac CT datasets: the Multi-Modality Whole Heart Segmentation (MM-WHS)~\cite{zhuang2019} and ASOCA~\cite{gharleghi2022} datasets for segmentation and classification stages of the pipeline, respectively. The MM-WHS dataset consists of 20 labelled CT volumes from healthy patients, each annotated with seven cardiac structures:  Myocardium, Left Atrium, Left Ventricle, Right Atrium, Right Ventricle, Aorta, Pulmonary Arteries. We both fine-tune Anatomix on this dataset and use it to train the registration component of the Atlas-ISTN. In our experiments, we split the MM-WHS dataset into training, validation, and testing sets in a 13/1/6 ratio, respectively.The uncertainty and segmentation error maps produced for the test set were used for both quantitative and qualitative evaluation. 

Downstream CVD classification utilizes the ASOCA dataset, which contains 40 coronary CT angiography scans, evenly split between healthy and diseased subjects. Due to the small dataset size, we use stratified 5-fold cross-validation (over three random seeds), and report results averaged across all runs. For the disease classification task, we employ a simple multilayer perceptron (MLP). The following hyperparameters were tuned using Bayesian Optimization \cite{akiba2019optuna} for each feature set: Hidden layer count \(L\in[1,12]\), Hidden unit size \(H\in[8,512]\), Dropout probability \(p\in[0,0.5]\), Learning rate \(\eta\in[10^{-4},10^{-2}]\) (log-uniform), Number of training epochs \(E\in\{100,125,150,\ldots,400\}\)

\subsection{Which uncertainty metric tracks segmentation error?}\label{seg-unc-analysis} 
We first ask which uncertainty representation best localises where the segmentation is erroneous. For each test subject in MM-WHS, we generate an \emph{error map}, which is the voxel-wise mismatch between the consensus Anatomix segmentation and the ground-truth. Next, we compute the entropy, KL, and aleatoric maps in both multi-class and per-structure form. We measure the normalised cross-correlation (NCC) between each uncertainty map and the error map.  NCC captures how strongly two maps co-vary spatially, with higher values indicating that uncertainty is high precisely where the segmentation is wrong. 

\begin{table*}[t]
\centering
\caption{Mean normalised cross correlation (NCC) between each uncertainty map and the segmentation error map on MM-WHS, per structure. Higher values indicate stronger correspondence between uncertainty and segmentation error.
}
\label{tab:ncc_mean}
\resizebox{\textwidth}{!}{%
\setlength{\tabcolsep}{7pt} 
\begin{tabular}{lcccccccc}
\toprule
&
&
&
\textbf{Left} &
\textbf{Left} &
\textbf{Right} &
\textbf{Right} &
&
\textbf{Pulmonary} \\
\textbf{Method} &
\textbf{Overall} &
\textbf{Myocardium} &
\textbf{Atrium} &
\textbf{Ventricle} &
\textbf{Atrium} &
\textbf{Ventricle} &
\textbf{Aorta} &
\textbf{Artery} \\
\midrule
Entropy (total) &
0.521 & 0.525 & 0.501 & 0.524 & 0.528 & 0.494 & 0.515 & 0.462 \\

KL (epistemic) &
0.408 & 0.362 & 0.308 & 0.433 & 0.366 & 0.401 & 0.392 & 0.330 \\

Aleatoric &
0.510 & 0.524 & 0.505 & 0.517 & 0.523 & 0.487 & 0.514 & 0.461 \\
\bottomrule
\end{tabular}
}
\end{table*}

\begin{figure}[t]
\centering
\includegraphics[width=0.97\textwidth]{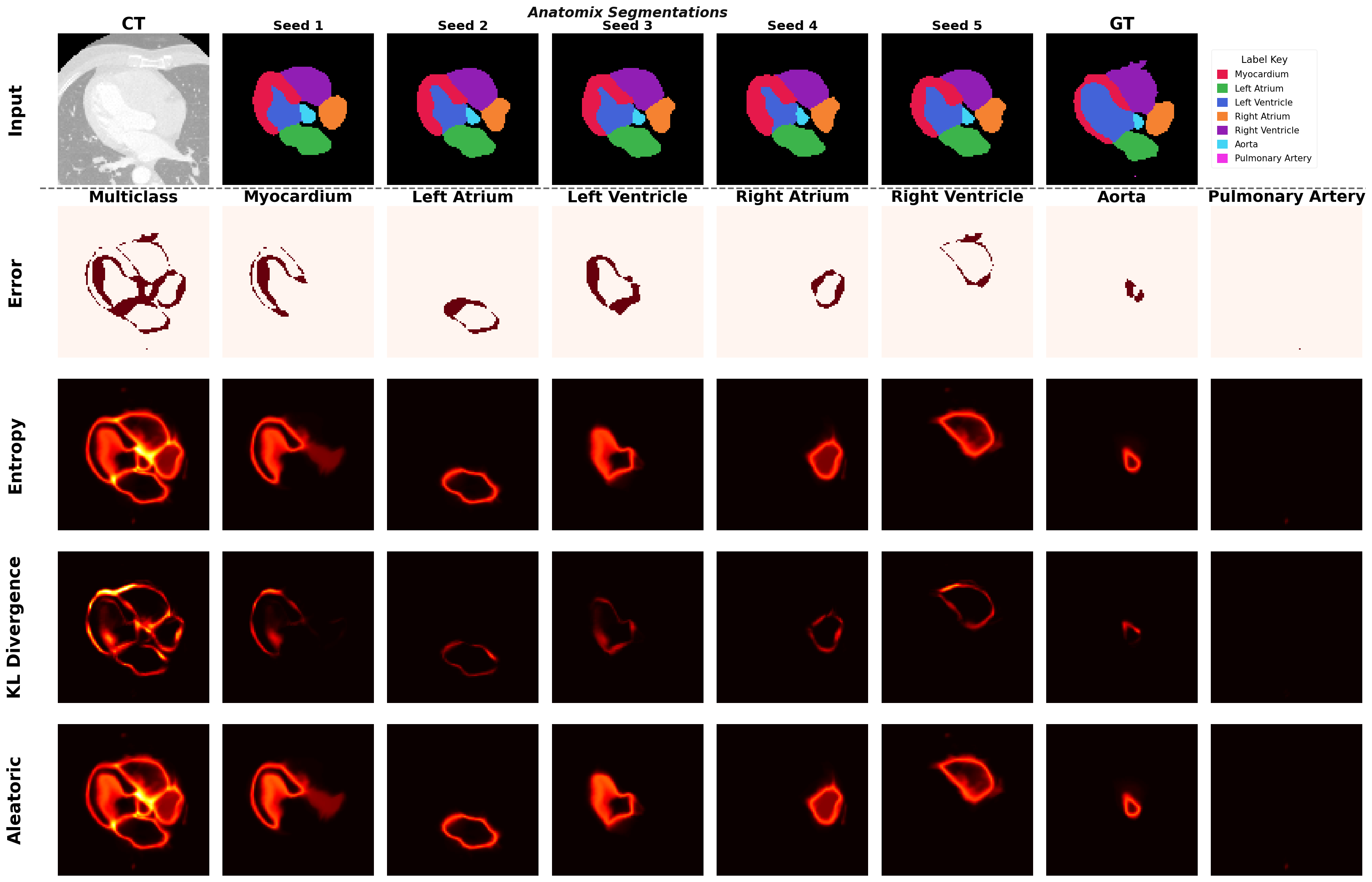}
\caption{Representative sample showing the original CT image, Anatomix segmentations, and ground-truth segmentation (top row), followed by the corresponding error maps and entropy, KL-divergence, and aleatoric uncertainty maps (multi-class and per-structure). Brighter regions indicate higher uncertainty.}
\label{fig:uncertainty_maps}
\end{figure}

Table~\ref{tab:ncc_mean} shows the total entropy achieves the highest correspondence with segmentation error (overall NCC $0.521$), the aleatoric map follows very closely ($0.510$), and the KL divergence trails substantially ($0.408$). The single-sample visualisation in Fig.~\ref{fig:uncertainty_maps} provides a qualitative evaluation. The entropy and aleatoric maps contain the most detail, and they co-localise with the error regions, whereas the KL map concentrates on sharp inter-seed disagreement at thin structure boundaries. The observation that the entropy and aleatoric maps behave almost identically, whereas the KL (epistemic) map is significantly weaker, suggests that segmentation errors are caused primarily by inherent ambiguity in the data. Consequently, one might expect total entropy and aleatoric uncertainty to be the most useful signals to propagate for downstream tasks. We test this hypothesis by measuring how classification performance changes when each uncertainty representation is incorporated into the feature descriptor.

\subsection{Which uncertainty encoding best improves classification?} The deterministic \texttt{GRC-Net} feature vector ($110 \times 7$ features) serves as our baseline, and we evaluate six uncertainty-augmented variants, each appending one uncertainty summary to the baseline vector, differing only in \emph{what} is appended: 

\begin{center}
\fbox{%
\begin{minipage}{0.97\linewidth}
\centering
\small
\textbf{Uncertainty-augmented variants (appended to the 770-dim \texttt{GRC-Net})}\\[5pt]

\footnotesize
\setlength{\tabcolsep}{4pt}

\begin{tabular}{p{0.48\linewidth} p{0.48\linewidth}}
(\texttt{E.G.}) Entropy, global ($+1$) & (\texttt{E.S.}) Entropy, per-structure ($+7$) \\
(\texttt{K.G.}) KL, global ($+1$) & (\texttt{K.S.}) KL, per-structure ($+7$) \\
(\texttt{A.G.}) Aleatoric, global ($+1$) & (\texttt{A.S.}) Aleatoric, per-structure ($+7$) \\
\end{tabular}

\end{minipage}}
\end{center}

\begin{table}[t]
\caption{Downstream CVD classification on ASOCA: deterministic baseline vs uncertainty-augmented variants. Mean{\tiny$\pm$std} over 5-fold CV and 3 seeds. Best results per metric are highlighted in \textbf{bold}.}
\label{tab:enc_results}
\centering
\small
\setlength{\tabcolsep}{3.5pt}
\renewcommand{\arraystretch}{1.1}

\begin{tabular}{llccc}
\toprule
& Method & AUROC (\%) & Precision (\%) & Recall (\%)  \\
\midrule

& \texttt{GRC-Net\cite{mittal2025}} & 91.25\tiny{$\pm$09.35} & 82.00\tiny{$\pm$13.96} &
90.00\tiny{$\pm$12.25} \\

\midrule

\multirow{6}{*}[-4pt]{\begin{turn}{90}\begin{tabular}{@{}c@{}}\texttt{GRC-ProbNet} \\ variants\end{tabular}  \end{turn}} & + \texttt{E.G}. & 89.58\tiny{$\pm$10.12} & 81.11\tiny{$\pm$12.50} &
\textbf{91.67}\tiny{$\pm$14.91}  \\

& + \texttt{E.S.} & 92.71\tiny{$\pm$10.68} & 84.59\tiny{$\pm$13.81} &
\textbf{91.67}\tiny{$\pm$14.91} \\

& + \texttt{K.G.} & 85.83\tiny{$\pm$13.05} & 83.33\tiny{$\pm$14.64} &
88.33\tiny{$\pm$15.46} \\

& + \texttt{K.S.} & \textbf{92.92}\tiny{$\pm$08.50} & \textbf{88.48}\tiny{$\pm$13.40} &
88.33\tiny{$\pm$12.47} \\

& + \texttt{A.G.} & 91.25\tiny{$\pm$10.72} & 83.44\tiny{$\pm$14.16} &
88.33\tiny{$\pm$15.46} \\

& + \texttt{A.S.} & 89.38\tiny{$\pm$09.67} & 82.00\tiny{$\pm$11.90} &
90.00\tiny{$\pm$15.28} \\

\bottomrule
\end{tabular}
\end{table}

\noindent
Global variants append a single scalar feature; per-structure variants append one feature per cardiac structure.

To ensure that observed differences can be attributed to the encoding rather than to training variability, all variants use the same ASOCA cross-validation folds and seed configurations. The \texttt{GRC-Net} baseline is also run under identical conditions. Results are reported in Table~\ref{tab:enc_results}. We observe that segmentation uncertainty does carry a diagnostically useful signal. Every per-structure variant (\texttt{E.S.}, \texttt{K.S.}, and \texttt{A.S.}) matches or exceeds the deterministic baseline in accuracy, and the best encodings (\texttt{K.S.} and \texttt{E.S.}) improve classification performance for almost all metrics. 

A clear pattern separates \emph{how} uncertainty is encoded from \emph{whether} it is included. Per-structure encodings (\texttt{E.S.}, \texttt{K.S.}, and \texttt{A.S.}) consistently outperform their global single-scalar counterparts (\texttt{E.G.}, \texttt{K.G.}, and \texttt{A.G.}). A single number captures only how uncertain the model is overall and discards \emph{where} that uncertainty lies. In contrast, per-structure features can preserve the spatial information necessary to detect region-specific failure modes. 

The most striking result is the relationship between these results and the error-localisation analysis in the previous section (Sec.~\ref{seg-unc-analysis}). Entropy showed the strongest correlation with segmentation error, while KL was the weakest by a clear margin. However, for downstream classification, KL matches entropy in AUC, and achieves slightly higher precision. This suggests that accurately identifying segmentation errors is not the same as identifying disease. Entropy is dominated by boundary ambiguity that occurs in both healthy and diseased cases, and this makes it a strong indicator of segmentation error but a less specific marker of disease. In contrast, KL captures inter-seed disagreement, which is a measure of epistemic uncertainty. Diseased structures are likely to lie closer to the segmentation model's decision boundary, and they may produce more of this disagreement than healthy anatomy. Overall, the uncertainty measure that best predicts segmentation error is not necessarily the one that provides the strongest signal for disease classification. 

\subsection{Feature Importance Analysis}

\begin{figure}[t]
\centering
\includegraphics[width=1\textwidth]{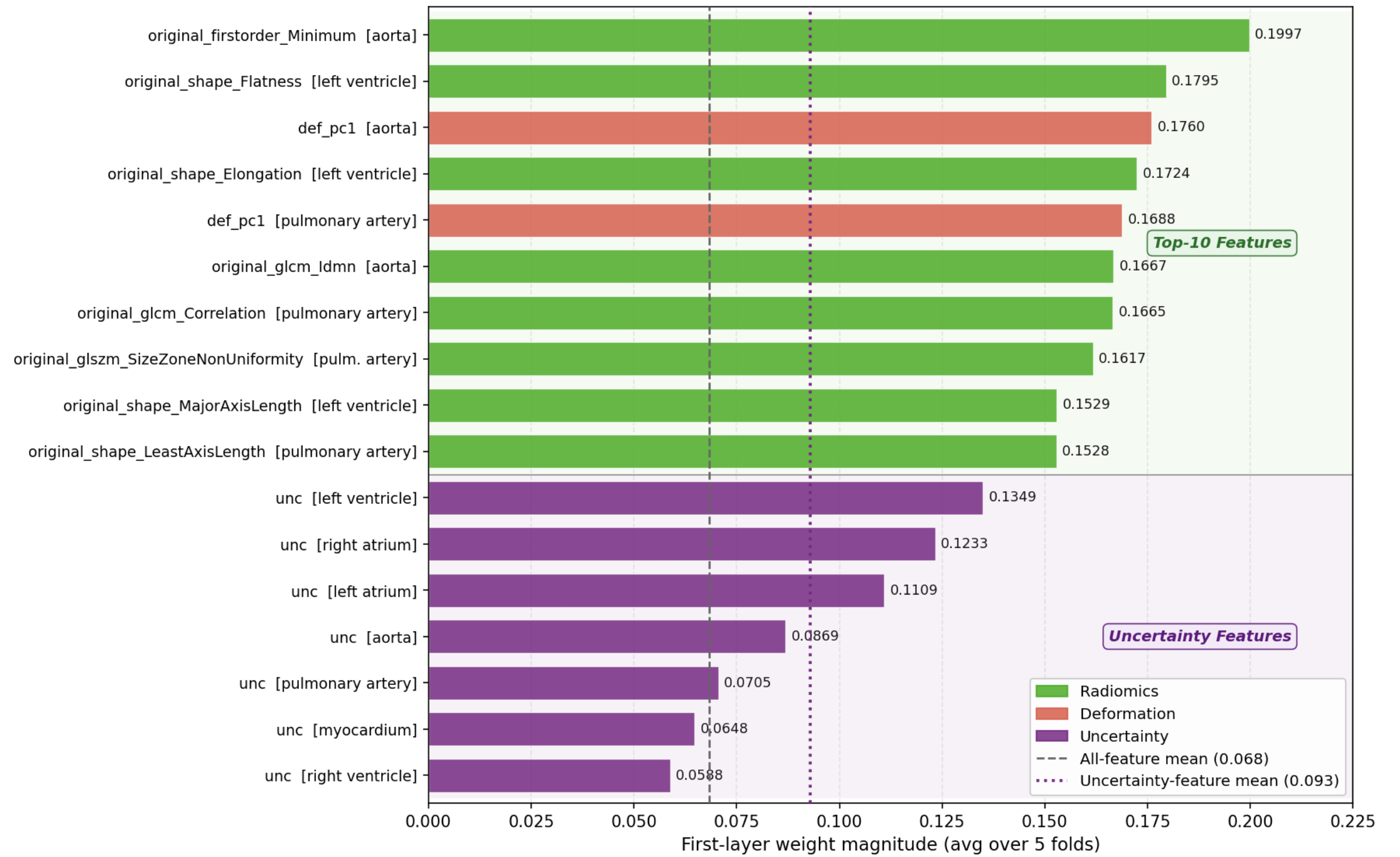}
\caption{First-layer weight magnitudes for the uncertainty-augmented classifier, grouped by feature type. Uncertainty-derived features are highlighted in purple to indicate their relative contribution compared to the top-10 radiomic (green) and geometric (orange) descriptors in the learned feature representation.}
\label{fig:feature-importance}
\end{figure}

To determine whether the classifier actively utilized the uncertainty inputs, a feature importance analysis was conducted using first-layer weight magnitudes. This metric quantifies the initial weight assigned to each input feature by the model, providing a straightforward indication of the inputs upon which the classifier predominantly relies.

The analysis revealed (Fig.~\ref{fig:feature-importance}) that, although the uncertainty features were not the highest-ranked overall, the majority received above-average weights relative to the global feature distribution. Specifically, the average weight magnitude across the uncertainty features was 0.0929, compared to a global feature mean of 0.0683. Averaged across all folds, these values provide a robust indication that the classifier learned to leverage the uncertainty inputs, despite the inclusion of only seven such features. Ultimately, this suggests that the entropy-derived features contribute genuine downstream signal rather than mere noise.

\section{Discussion and Conclusion}

This work investigated whether segmentation uncertainty can improve a modular hybrid pipeline for cardiovascular disease classification. Our results show that uncertainty provides complementary information beyond deterministic segmentations, with \texttt{GRC-ProbNet} improving classification performance over the baseline \texttt{GRC-Net} pipeline. More importantly, the choice of uncertainty representation has a clear effect on downstream performance, showing that how uncertainty is encoded is as important as including uncertainty itself. A single global uncertainty number tends to lose spatial information and is not as useful for classification compared to structure-specific uncertainty. These findings also suggest that evaluating uncertainty solely by its ability to predict segmentation error may be insufficient. The representation that best reflects segmentation quality is not necessarily the one that provides the strongest signal for downstream disease classification, and vice versa. This is crucial in many medical imaging pipelines that involve both deep- and non-deep-learning components, since uncertainty is explicitly represented before feature extraction and classification, rather than being absorbed into an end-to-end learned model. Future work will evaluate the framework on larger cohorts, incorporate multiple expert annotators directly into the pipeline to model the gold standard of inter-observer variability, and finally, investigate uncertainty propagation through the registration stage.

\section*{Acknowledgment}
This work was supported by the European Union's Horizon Europe research and innovation programme for the AI-POD project under grant agreement \texttt{101080302}. Views and opinions expressed are however those of the author(s) only and do not necessarily reflect those of the European Union or HaDEA. Neither the European Union nor the granting authority can be held responsible for them.

\subsection*{Disclosure of interests}
B.G. is a part-time employee of DeepHealth. No other competing interests.

\end{document}